\documentclass[runningheads]{llncs}

\usepackage{graphicx}
\usepackage{amsmath,amssymb} 
\usepackage{color}
\usepackage[width=122mm,left=12mm,paperwidth=146mm,height=193mm,top=12mm,paperheight=217mm]{geometry}

\usepackage{courier}
\usepackage{cite}

\usepackage[colorlinks,linkcolor = red ,anchorcolor = blue ,citecolor = green]{hyperref}

\begin{document}

\pagestyle{headings}
\mainmatter
\def\ECCV18SubNumber{***}  

\title{Adversarial 3D Human Pose Estimation via Multimodal Depth Supervision}
\authorrunning{K. Zhou \emph{et al.}}
\author{Kun Zhou$^{1}$, Jinmiao Cai$^{2,3}$, Yao Li$^{2,3}$, Yulong Shi$^{3}$ \newline
Xiaoguang Han$^{4,3}$, Nianjuan Jiang$^{1,3}$, Kui Jia$^{2}$, Jiangbo Lu$^{1,3}$
}

\institute
{
$^{1}$Shenzhen Cloudream Technology Co., Ltd. ~~ $^{2}$South China University of Technology \newline
$^{3}$Cloudream AI3D Lab ~~ $^{4}$Chinese University of Hong Kong~(Shenzhen) \newline
}

\maketitle
\vspace{-15pt}
\begin{abstract}
In this paper, a novel deep-learning based framework is proposed to infer 3D human poses from a single image. Specifically, a two-phase approach is developed. We firstly utilize a generator with two branches for the extraction of explicit and implicit depth information respectively. During the training process, an adversarial scheme is also employed to further improve the performance. The implicit and explicit depth information with the  estimated 2D joints generated by a widely used estimator, in the second step, are together fed into a deep 3D pose regressor for the final pose generation. Our method achieves MPJPE of 58.68mm on the ECCV2018 3D Human Pose Estimation Challenge.

\end{abstract}

\vspace{-15pt}
\section{Introduction}
Estimating the 3D human pose from a single RGB image~\cite{IonescuSminchisescu11, pishchulin2016deepcut, bogo2016keep, shi2018fbi, pavlakos2018ordinal, zanfir2018monocular, marinoiu20183d,zhou2017towards} has drawn intensive research attentions over the last decade due to its broad applications. Thanks to the powerful DCNN~(deep convolutional neural network), significant advances have been witnessed in this area. Nevertheless, there still exists a large gap between images and 3D poses for in-the-wild scenarios. This occurs as the result of the challenges to annotate 3D groundtruth positions for skeleton joints.

Many of previous works tackle this problem by decomposing the task into two stages, each aiming at training models with easy annotating: (a)~performing 2D pose estimation; (b)~recovering 3D pose from 2D pose directly. Although this decomposition helps with data annotations, as 2D pose annotation can be more easily obtained with in-the-wild images, it also discards necessary pictorial information for resolving the ambiguity in 3D pose recovery. Depth ordering information e.g.,~\cite{shi2018fbi, pavlakos2018ordinal}, has been demonstrated to be effective for solving this ambiguity and the annotation can be done efficiently.

Taking~\cite{shi2018fbi} as consideration, the \emph{FBI~(forward-or-backward information)} only partially reflects the absolute depth information. To further advance along this line, we propose an architecture for extracting multimodal depth information. More specifically, both the FBI, as an implicit depth information, and the explicit depth information are exploited to supervise the learning procedure. In addition, we improve the estimation of explicit depth by using a conditional adversarial learning scheme. At the last stage, a linear deep regressor with a novel loss function maps FBI and the explicit depth with the corresponding 2D human pose into the estimated 3D human pose.



\begin{figure}[t]
\centering
\includegraphics[height=5cm]{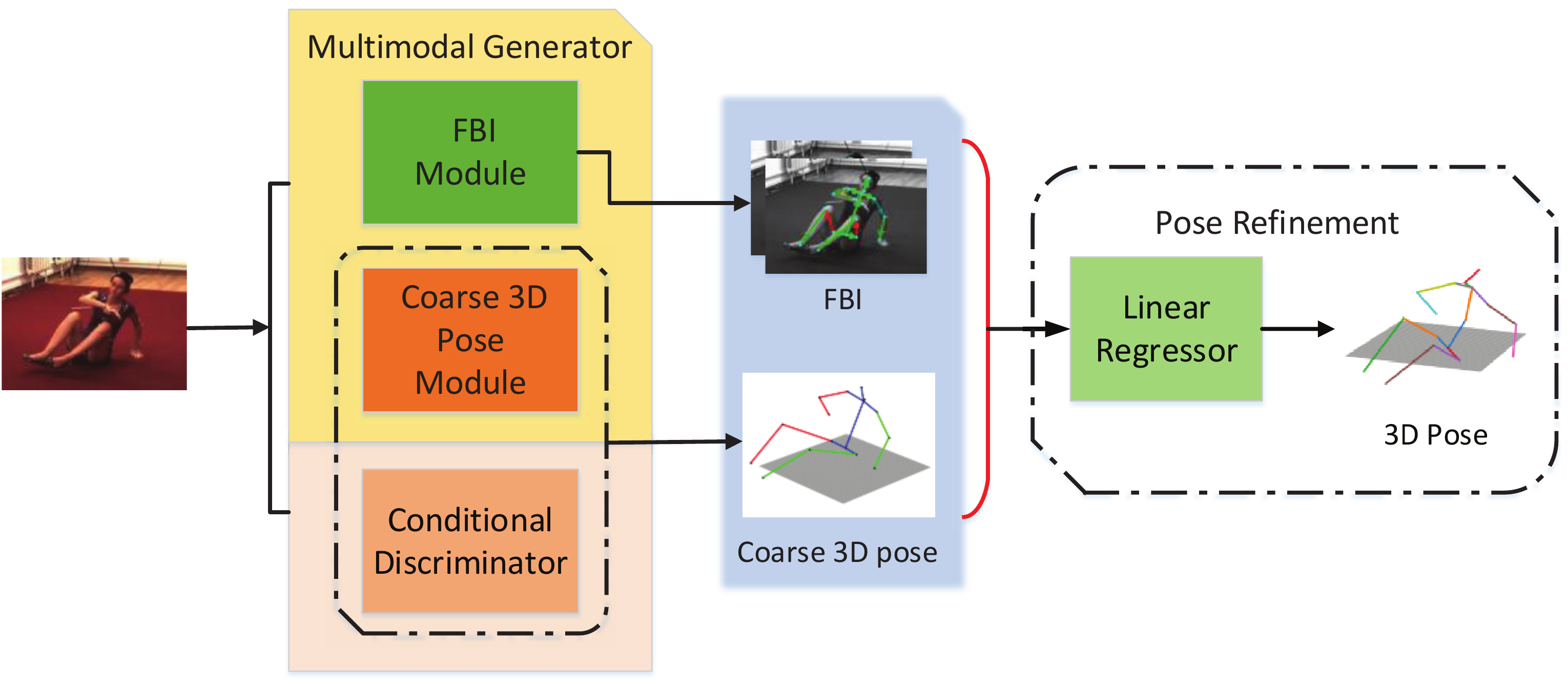}
\caption{The framework of our method. A multimodal generator is learned to predict a coarse 3D pose and FBI in a supervised way. An adversarial training is adopted to boost the performance of the part for 3D pose inference. Both the estimated coarse 3D pose and FBI are then together fed into a deep regressor for further 3D pose reﬁnement.}

\label{fig:overview}
\end{figure}

\vspace{-10pt}
\section{Method}
\vspace{-5pt}

The overview of the proposed network architecture is shown in Fig.~\ref{fig:overview}. A coarse 3D pose and FBI are estimated by our multimodal generator. A conditional adversarial learning architecture is employed to fine-tune the coarse 3D pose module. Finally, the coarse 3D pose and FBI are fed into the linear regressor to infer the 3D human pose. 

\vspace{-13pt}
\subsection{Multimodal Depth Estimation}
\vspace{-3pt}

\subsubsection{Multimodal Generator.}
The multimodal generator consists of two parallel convNets. One convNet estimates a coarse 3D pose, an explicit depth representation, and the other convNet generates FBI, an implicit depth representation. 

\vspace{-13pt}
\subsubsection{Explicit Depth Supervision.}
Our explicit depth is represented by assigning an extra $z$ coordinate on each 2D joint location. In this work, the 2D joint locations are detected by a widely-used estimator\cite{newell2016stacked}. The explicit depth with the corresponding 2D joint locations can be viewed as a coarse 3D pose aligned in the camera coordinate system. Such information can be easily extracted from the 3D pose groundtruth, and used to supervise the learning process. We use the same convNets architecture as in~\cite{zhou2017towards} for our coarse 3D pose prediction.

\vspace{-13pt}
\subsubsection{Implicit Depth Supervision.}
FBI is an implicit depth information indicating if a bone is forward-or-backward facing with respect to the camera's view. We selected $m~(=~14)$ FBI relationships from a human skeleton. Each bone vector $\overrightarrow B_i~(i=1,2,..,m)$ has one of three status: definitely forward, definitely backward and possibly parallel to the image plane. The FBI of an image can be defined as a matrix \emph{F} = $\{f_1, f_2, ..., f_m\}$ where $f_i$ is a one-hot 3-dimensional vector, i.e., $f_i (j)=1~(j =0,1,2)$ means the $i_{th}$ bone has the $j_{th}$ status. Interested readers are referred to \cite{shi2018fbi} for more details of FBI. It is worth noting that such information is very easy to annotate. The users only need to do a binary selection for each skeleton bone, incurring around 20 seconds for each image. We use the same convNets as in \cite{shi2018fbi} for FBI estimation.

\begin{figure}[t]
\centering
\includegraphics[height=4cm]{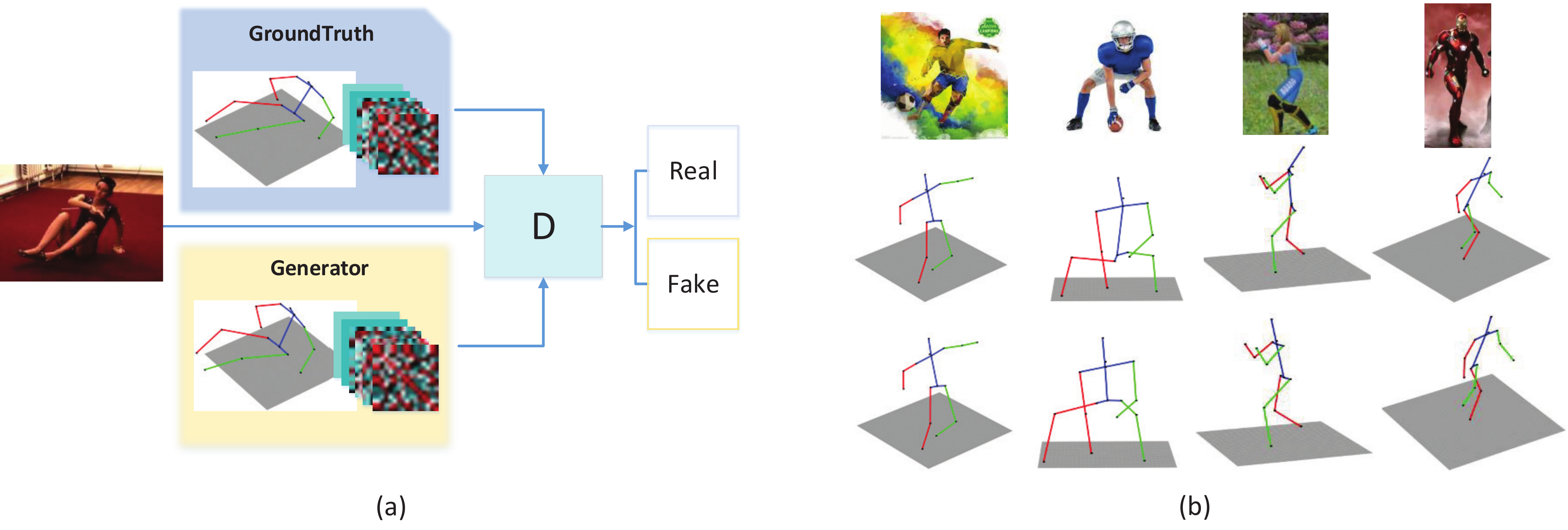}
\caption{(a) The architecture of conditional adversarial learning. Discriminator training is used to distinguish the authenticity of the samples and the generator is used to generate 3D poses that are anthropometrically valid to fool the discriminator. (b) Visual results of estimated coarse 3D poses for out-of-the-domain images. Second/Third row: results without/with conditional adversarial learning.
}
\label{fig:gan-discriminator}
\end{figure}

\vspace{-15pt}
\subsubsection{Adversarial Learning.} 
It has been proved in~\cite{yang20183d} that adversarial learning helps to predict more realistic poses. The whole process of our conditional adversarial learning is depicted in Fig.~\ref{fig:gan-discriminator}(a). In the pre-training stage, a coarse 3D pose is predicted by our generator. Subsequently, the coarse 3D pose is refined by conditional adversarial learning in the fine-tuning stage. Real and fake labels are generated from the discriminator, which in turn leads to generating plausible 3D poses. Thanks to the generalization power of conditional adversarial learning, our coarse 3D module is robust. The visual results of some images with very different domain characteristics are shown in Fig.~\ref{fig:gan-discriminator}(b).

\vspace{-13pt}
\subsection{3D Pose Refinement}
\vspace{-3pt}
Multimodal depth features fed into the network can effectively improve the accuracy of 3D pose estimation. Specifically, the coarse 3D pose and FBI are concatenated together and then mapped into the 3D pose by exploiting two cascaded linear regression blocks used in~\cite{martinez2017simple}.

\vspace{-13pt}
\subsubsection{Weighted Regression Loss function.}
Let \emph{p} be the predicted 3D pose and \emph{P} be the groundtruth 3D pose. The loss function for 3D pose regression is:

\begin{equation}
Loss(p) = L^0 + \varepsilon L^1
\label{regression loss}
\end{equation}

\vspace{-10pt}
\begin{equation}
L^1 = (L^0+\alpha)^2 L^0 e^{L^0/(1-\alpha)}
\label{L1}
\end{equation}

\vspace{-1pt}
\noindent Here, $L^0$ is the basic L2 loss. $1-\alpha$ is the mean of $L^0$ on the training dataset. $\varepsilon$ is a hyperparameter to adjust the trade-off between $L^0$ and $L^1$, and is set as 0.001 in the experiments. Actions with large poses are commonly hard to learn, and these hard samples should gain more attention. To this end, $L^1$ is designed to complement $L^0$ which allows different samples to get adaptive supervision focus.

\vspace{-10pt}
\section{Experiments}
\vspace{-5pt}

\subsection{Training}
\vspace{-3pt}
\subsubsection{Dataset.}
Only 3D human pose data from the ECCV Challenge dataset, a subset of the large-scale dataset Human3.6M\cite{IonescuSminchisescu11, h36m_pami}, is used in the training process. FBI and our coarse 3D pose derived from the ECCV Challenge dataset are used for our multimodal generator training.

\vspace{-13pt}
\subsubsection{Implementation Details.}
The whole framework is implemented on Tensorflow. The linear 3D pose regressor requires less than six hours for training.

The ECCV 3D Pose Challenge only provides RGB images and the corresponding 3D pose coordinates groundtruth. We use the 2D pose estimator~\cite{newell2016stacked} to assist in cropping the full human body in an image and resize it to 256$\times$256.

\vspace{-13pt}
\subsubsection{Results.}
All the results in Table~\ref{table:results} are obtained from the evaluation server. The method inferring 3D poses only from 2D joint locations without the coarse 3D pose and FBI is denoted as "Base". The full method is denoted as "Final".

\vspace{-10pt}
\section{Analysis}
\vspace{-5pt}
It is clear that the most challenging task for 3D human pose estimation is the learning of depth. We proposed to simultaneously infer the explicit depth and implicit depth, in a supervised manner, using a convNets architecture with two independent branches. Despite the FBI lacks groundtruth of explicit depth for in-the-wild images, it can provide  useful depth supervision,  and it is also very easy to annotate. We take complementary advantages of the implicit and explicit depth supervision and feed the learned features together to the final regressor for 3D pose inference. A weighted regression loss function provides an adaptive feedback for different pose samples. Thanks to these designs, our proposed method achieves competitive 3D human pose estimation.

\setlength{\tabcolsep}{4pt}
\begin{table}[t]\scriptsize
\begin{center}
\caption{Results on the official evaluation server~(measured in millimeter).}
\vspace{-10pt}
\label{table:results}
\begin{tabular}{l|lllllllllllllll|l}
\hline\noalign{\smallskip}
MPJPE & A1 & A2 & A3 & A4 & A5 & A6 & A7 & A8 & A9 & A10 & A11 & A12& A13 & A14 & A15 & Avg\\
\noalign{\smallskip}
\hline
\noalign{\smallskip}
Base  & 54 & 56 & 55 & 53 & 59 & 55 & 69 & 74 & 89 & 63 & 61 & 60 & 53 & 65 & 59 & 62\\
Final  & {\bf 53} & {\bf 54} & {\bf 54} & {\bf 52} & {\bf 56} & {\bf 55} & {\bf 58} & {\bf 70} & {\bf 78} & {\bf 60} & {\bf 59} & {\bf 57} & {\bf 48} & {\bf 61} & {\bf 56} & {\bf 58.68}\\
\hline
\end{tabular}
\end{center}
\end{table}
\setlength{\tabcolsep}{1.4pt}

We also find that the 2D joint coordinates of our coarse 3D pose is not reliable enough (left-right joint pairs sometimes flip). As future work, we will explore combining a stronger 2D pose detector and a more effective depth feature extractor for 3D human pose estimation.

\clearpage
\bibliographystyle{splncs}
\bibliography{abstract}
\end{document}